\documentclass[letterpaper, 10 pt, conference]{ieeeconf}  

\IEEEoverridecommandlockouts                              

\usepackage{graphics} 
\usepackage{graphicx}
\usepackage{float}
\usepackage{times} 
\usepackage{amsmath} 

\usepackage{algorithm}
\usepackage{algorithmic}

\usepackage{subcaption}
\usepackage{tabulary,booktabs}
\usepackage{multirow}
\usepackage{multicol}
\usepackage{threeparttable}
\usepackage{color}
\usepackage{hyperref}

\title{\LARGE \bf
Unsupervised Visible-light Images Guided\\ Cross-Spectrum Depth Estimation from Dual-Modality Cameras
}

\author{
Yubin Guo$^{\dag}$,
Haobo Jiang$^{\dag}$, 
Xinlei Qi$^{\dag}$,
Jin Xie$^{\dag}$, 
Cheng-Zhong Xu$^{*}$,
Hui Kong$^{*}$
\thanks{$\dag$ School of Computer Science and Engineering, Nanjing University of Science and Technology, Nanjing, Jiangsu, China.}
 \thanks{$\ddag$ The State Key Laboratory of Internet of Things for Smart City (SKL-IOTSC), Department of Computer Science, University of Macau, Macau, China.}
 \thanks{$*$ The State Key Laboratory of Internet of Things for Smart City (SKL-IOTSC), Department of Electromechanical Engineering (EME), University of Macau, Macau, China.}
}

\begin{document}

\maketitle
\thispagestyle{empty}
\pagestyle{empty}

\begin{abstract}
Cross-spectrum depth estimation aims to provide a depth map in all illumination conditions with a pair of dual-spectrum images. It is valuable for autonomous vehicle applications when the vehicle is equipped with two cameras of different modalities. However, images captured by different-modality cameras can be photometrically quite different. Therefore, cross-spectrum depth estimation is a very challenging problem. Moreover, the shortage of large-scale open-source datasets also retards further research in this field. In this paper, we propose an unsupervised visible-light image guided cross-spectrum (i.e., thermal and visible-light, TIR-VIS in short) depth estimation framework given a pair of RGB and thermal images captured from a visible-light camera and a thermal one. 
We first adopt a base depth estimation network using RGB-image pairs. Then we propose a multi-scale feature transfer network to transfer features from the TIR-VIS domain to the VIS domain at the feature level to fit the trained depth estimation network. At last, we propose a cross-spectrum depth cycle consistency to improve the depth result of dual-spectrum image pairs. Meanwhile, we release a large dual-spectrum depth estimation dataset with visible-light and far-infrared stereo images captured in different scenes to the society. The experiment result shows that our method achieves better performance than the compared existing methods. Our datasets is available at  \url{https://github.com/whitecrow1027/VIS-TIR-Datasets}.

\end{abstract}

\section{Introduction}

The application of the dual-spectrum visual system is gradually increasing with the development of computer vision. Images captured by a thermal infrared (TIR) camera are related to object thermal radiation and not affected by illumination conditions as much as images by visible-light (VIS) cameras. Compared to the vision system with only visible-light cameras, Dual-spectrum visual system with additional TIR cameras is more robust to different illumination conditions such as dark or overexposed scenes. Up to date, many researchers have combined cameras of different spectra for various computer vision tasks, such as pedestrian detection \cite{liu2016multispectral} \cite{li2019illumination}, person re-identification \cite{wu2017rgb}, tracking \cite{zhu2020quality}, and SLAM \cite{dai2019multi} \cite{shin2019sparse}.

\begin{figure*}[ht]
    \centering
    \begin{subfigure}[b]{1.7in}
        \includegraphics[width=1.7in]{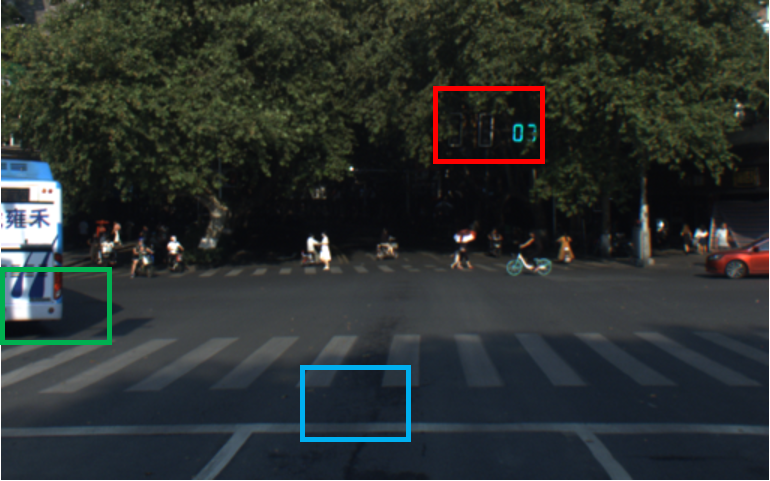}
        \caption{True VIS}
    \end{subfigure}
    \begin{subfigure}[b]{1.7in}
        \includegraphics[width=1.7in]{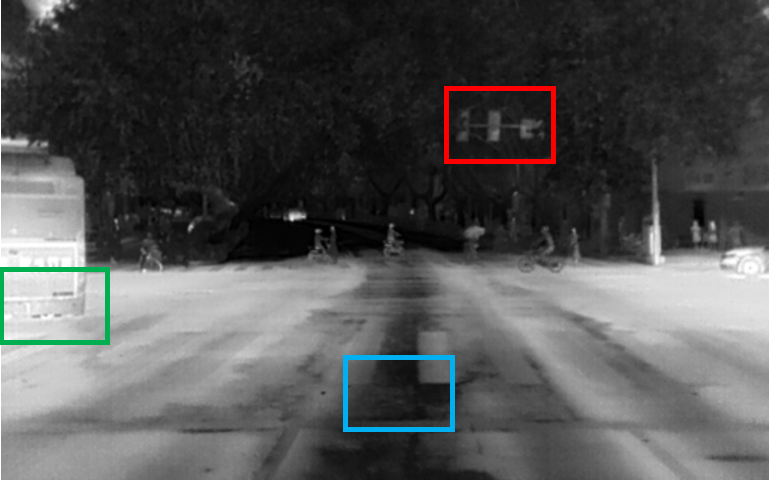}
        \caption{True TIR}
    \end{subfigure}
    \begin{subfigure}[b]{1.7in}
        \includegraphics[width=1.7in]{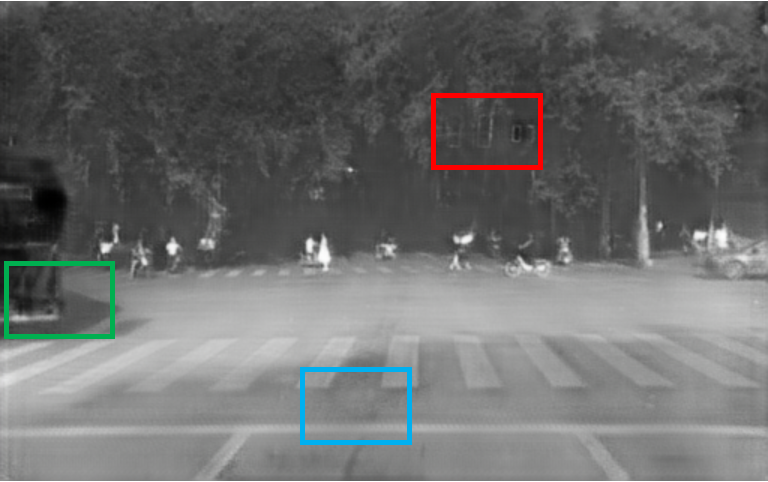}
        \caption{Synthesized TIR}
    \end{subfigure}
    \begin{subfigure}[b]{1.7in}
        \includegraphics[width=1.7in]{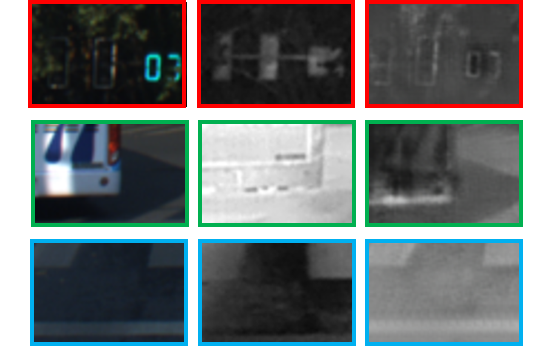}
        \caption{Texture Difference}
    \end{subfigure}
    \caption{Appearance variation between VIS, TIR and synthesized TIR image by cycleGAN \cite{zhu2017unpaired}. The synthesized TIR image looks real but has a large gap in texture compared with real TIR image due to environment temperature diversity.}
    \label{fig:fakeimg}
\end{figure*}

As a fundamental vision task, depth estimation aims to predict depth value for each pixel in a stereo or a single image. Now days, since getting per-pixel ground-truth depth for images based on LiDAR or structured-light sensors is very expensive and troublesome, many efforts have been invested on unsupervised methods for the depth estimation problem. In general, most unsupervised depth estimation methods based on visible-light cameras (with RGB images) \cite{garg2016unsupervised} \cite{zhou2017unsupervisedstereo} are mainly based on photometric consistency criterion. However, it is hard to apply this criterion to cross-spectrum images due to very significant variation between images of different modalities. 

To address this issue, Li \cite{li2020ivfusenet} uses a supervised method to train the dual-spectral depth estimation network with fusion TIR-VIS images. Kim \cite{kim2018multispectral} presents an unsupervised multi-spectral depth estimation method with pixel-aligned VIS images. Zhi \cite{zhi2018deep}  and Liang \cite{liang2019unsupervised} give a semi/unsupervised network to estimate depth for VIS-NIR(near-infrared) image pair by spectral transfer network, respectively. The two methods are designed for VIS-NIR images but not suitable for TIR-VIS images due to a larger spectral difference between TIR and VIS images. 

We observe some common issues in the existing depth estimation methods based on cross-spectrum images. (1) It is hard to find an effective self-supervision mechanism between images of different spectrum. (2) Some methods such as \cite{zhi2018deep} \cite{kim2018multispectral} \cite{liang2019unsupervised} pay their attention on image transferring between different spectrum, e.g., the images captured by thermal and visible-light cameras. Because the imaging mechanisms are different for the two modalities, the transformation is changeable due to environment temperature diversity.
Simply transferring images from one modality to another at the pixel level is not robust for depth estimation problem based on cross-spectrum images, as shown in Figure \ref{fig:fakeimg}. (3) There are barely open-source large-scale dual-spectrum datasets for the depth estimation problems, and the lack of such data retards further research in this direction.

In this work,  we propose an unsupervised visible-light image guided cross-spectrum depth estimation method given a pair of TIR-VIS images. 
The proposed framework is shown in Figure \ref{fig:fullnet}. 
First, we adopt an encoder-decoder depth estimation network, and train this network with VIS stereo image pairs. Then we propose a feature transfer network to transfer the feature from the TIR-VIS image pair by adversarial learning to fit the pre-trained depth decoder network. At last, we propose a depth-cycle consistency between different spectral images to further improve the depth estimation results. Meanwhile, we release a large-scale VIS-TIR dataset for the urban-road scene with depth ground truth which is obtained based on a multiple-channel LiDAR sensor. This dataset contains images that were captured under various illumination conditions and in different scenes. 



\section{Related Work}

\subsection{Unsupervised Depth Estimation}
With the development of convolutional neural network (CNN), Garg et al. first \cite{garg2016unsupervised} proposed a CNN method in an unsupervised way to estimate depth based on image reconstruction error loss between a calibrated stereo camera image pair. Based on this work, Zhuo \cite{zhou2017unsupervisedvideo} proposed a method to estimate depth from monocular video by synthesizing images between neighborhood frames with depth and camera pose. Godard \cite{godard2017unsupervised} \cite{godard2019digging} improved the monocular depth estimation result by adding a left-right consistency loss and extended his work to monocular and stereo cameras for both depth and pose estimation. To deal with the problem caused by moving objects in depth estimation, some works \cite{zou2018df} \cite{yin2018geonet}\cite{ranjan2019competitive} \cite{jiao2018look} \cite{bian2019unsupervised} additionally introduced optical flow or motion segmentation network to the depth estimation framework.
Zhou \cite{zhou2017unsupervisedstereo} proposed an unsupervised stereo matching method with an iterative disparity confidence map from the left-right reconstruction image. Sometimes, other vision tasks such as semantic segmentation \cite{chen2019towards} are combined into the depth estimation network to improve depth estimation. 

\begin{figure*}[ht]
    \centering
    \includegraphics[width=\textwidth]{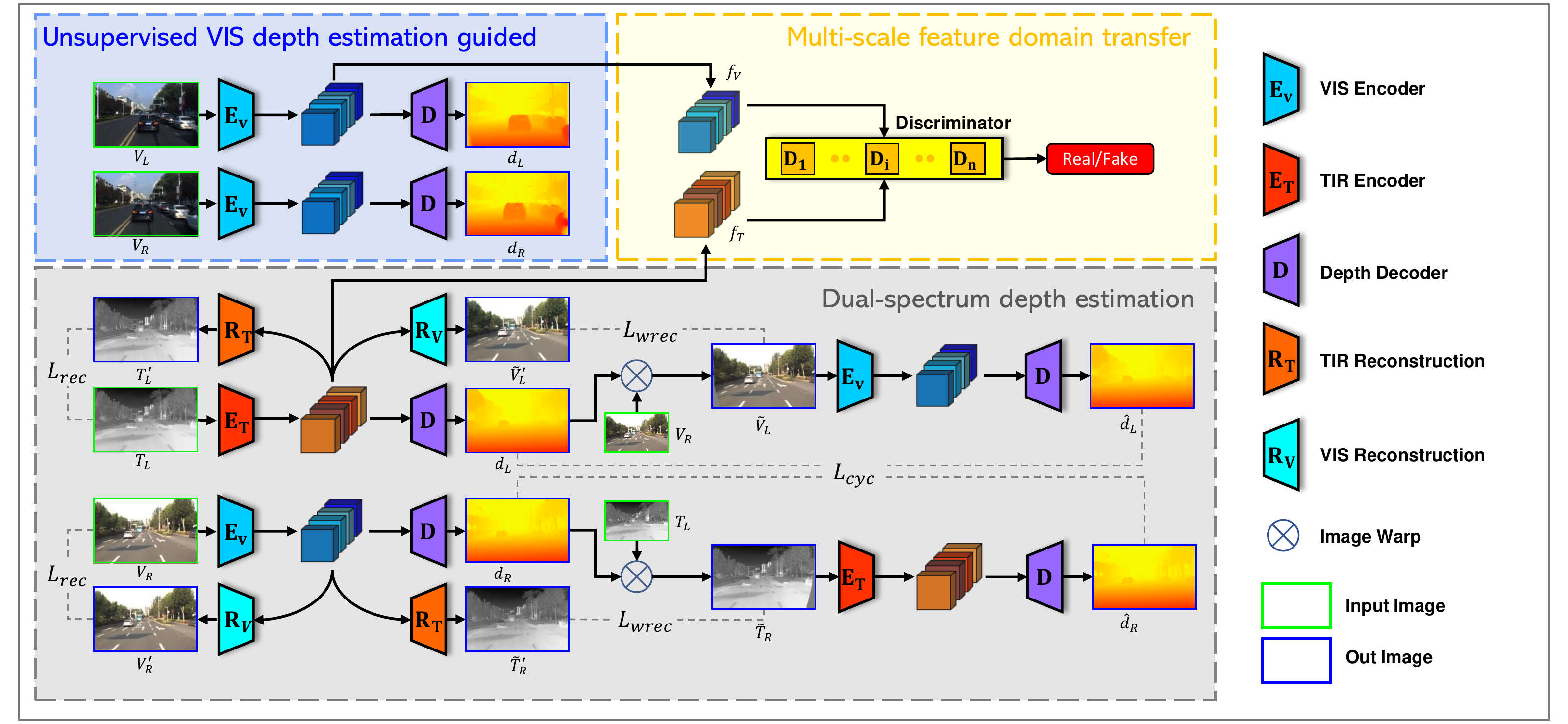}
    \caption{The pipeline of our VIS-guided cross-spectrum depth estimation framework. Our framework mainly contains three parts: (a) Unsupervised VIS depth estimation: an encoder-decoder depth estimation network for a pair of VIS images.  (b) Multi-scale feature domain transfer: an adversarial learning network that transfers TIR features to the VIS domain. (c) cross-spectrum depth estimation: given the TIR-VIS image pair $T_L,V_R$, extracting their feature to get depth output with depth decoder trained in (a) by combining of depth cycle consistency. All networks with the same symbol share parameters. }
    \label{fig:fullnet}
\end{figure*}

\subsection{Cross-spectrum Depth Estimation}
In recent years, more and more researchers start to consider the cross-spectrum depth estimation problem. Heo et al. \cite{heo2010robust} proposed a radiometric insensitive measurement called Adaptive Normalized Cross-Correlation (ANCC) for image matching with different illumination conditions. Aguilera et al. \cite{aguilera2012multispectral} propose an Edge Oriented Histogram (EOH) descriptor for matching interest points across different spectrum by utilizing edge information.
Kim et al. \cite{kim2015dasc} proposed an operator to find correspondence between images of different modalities by measuring local similarity based on hand-craft features. Zhi et al. \cite{zhi2018deep} built a visible-near infrared(VIS-NIR) stereo dataset for cross-spectrum depth estimation and proposed a VIS-NIR stereo image depth estimation method with the supervision of material labels. Liang \cite{liang2019unsupervised} used the cycleGAN-based image transfer model \cite{zhu2017unpaired} to synthesize images between VIS-NIR images to solve an unsupervised depth estimation problem based on the dataset in \cite{zhi2018deep}. Kim proposed a spectrum transfer model for cross-spectrum depth estimation with aligned VIS-TIR image pairs \cite{kim2018multispectral}. Li \cite{li2020ivfusenet} proposed a supervised model to fuse VIS-TIR features for depth estimation. 

\section{Method}
Our depth estimation framework is illustrated in Figure \ref{fig:fullnet}. Since TIR and VIS cameras share similar projective imaging model, images taken by TIR and VIS cameras for the same scene should have the same depth after calibrated. 
Our framework can be described in three steps. First, we train an unsupervised depth estimation network on VIS stereo image pairs. Then, we transfer TIR-VIS depth features to the VIS stereo features domain. At last, we get the depth estimation result from the transferred feature guided by the trained VIS depth estimation network and improve the result with cross-spectral depth consistency.

\subsection{Unsupervised VIS Stereo Depth Estimation}

Unsupervised VIS stereo depth estimation aims to estimate pixel-wise depth from a given stereo image pair without ground truth. We adopt the Encoder-Decoder network structure proposed in \cite{godard2017unsupervised}. Given a VIS image pair $(\boldsymbol{I}^{l},\boldsymbol{I}^{r})$, the VIS encoder $E_V$ encodes them to multi-scale features $\boldsymbol{f}^{l}$ and $\boldsymbol{f}^{r}$. The decoder $D$ decodes features to depth $\boldsymbol{d}^{l}$ and $\boldsymbol{d}^{r}$. Network $E_V$ and $D$ are connected by skip-connection as in U-net. We apply the photometric re-projection loss $L_{p}$, depth smoothness loss $L_{s}$ and left-right depth consistency loss $L_{lr}$.
The right-view image $(\boldsymbol{I}^{r})$ is re-projected to the left-view image $\widetilde{\boldsymbol{I}}^{l}$ with left-view depth $\boldsymbol{d}^{l}$ by camera intrinsic/extrinsic parameters,
$\widetilde{\boldsymbol{I}}^{l}=\omega(\boldsymbol{I}^{r},\boldsymbol{d}^{l})$.
Obviously, an accurate depth map should enable the re-projected image closer to the original image. We use \textit{ssim} and \textit{L1} distance to measure image photometric similarity. Thus, the photometric re-projection loss $L_{p}$ can be described as
\begin{align}
    \begin{split}
        L_{p}(\boldsymbol{I}^{l})= & \frac{1}{n}\sum_{i,j}\alpha \frac{1-ssim(\widetilde{\boldsymbol{I}}^{l}_{i,j},\boldsymbol{I}^{l}_{i,j})}{2} \\
        & +(1-\alpha)|| \widetilde{\boldsymbol{I}}^{l}_{i,j}-\boldsymbol{I}^{l}_{i,j}||_{1}
    \end{split}
\end{align}
where $ssim$ is simplified with a 3x3 block filter and $\alpha$ is set to 0.85.

Depth smoothness loss encourages depth to have smooth gradient locally. Left-right depth consistency loss ensures that the left-view depth to be equal to the re-projected right-view depth map to regularize the consistency of the left- and right-view depth:
\begin{align} \label{eq1}
    L_{s}(\boldsymbol{I}^{l})=|\partial x\boldsymbol{d}^{l}|e^{-||\partial x\boldsymbol{I}^{l}||_{1}}+|\partial y\boldsymbol{d}^{l}|e^{-||\partial y\boldsymbol{I}^{l}||_{1}}
\end{align}
\begin{align}
    L_{lr}(\boldsymbol{I}^{l})=||\boldsymbol{d}^{l}-\omega (\boldsymbol{d}^{r},\boldsymbol{d}^{l})||_{1}
\end{align}
where $\partial \boldsymbol{d}^{l}$ and $\partial \boldsymbol{I}^{l}$ means the depth and input image gradients.Thus, we get the total VIS unsupervised depth estimation loss
\begin{align} 
    \begin{split}
        L_{v} = & L_{p}(\boldsymbol{I}^{l})+L_{p}(\boldsymbol{I}^{r})+\lambda _{s}(L_{s}(\boldsymbol{I}^{l})+L_{s}(\boldsymbol{I}^{r})) \\
               & +\lambda _{lr}(L_{lr}(\boldsymbol{I}^{l})+L_{lr}(\boldsymbol{I}^{r}))
    \end{split}
\end{align}

\subsection{VIS-guided Cross-Spectrum Depth Estimation}
Inspired by \cite{goodfellow2014generative}, we adopt an unsupervised adversarial training to transfer features from the TIR domain to the VIS domain. According to \cite{liu2017unsupervised}, the feature level domain transfer could get both semantically similar information and domain-specific feature representation robustly. Thus, we pay more attention to domain transfer on the feature level instead of transferring images directly.

For ease of explanation, we denote VIS stereo image pair as source domain $\textbf{\textit{S}}$ and TIR-VIS stereo image pair as target domain $\textbf{\textit{T}}$. With a VIS depth estimation network trained with images in source domain $\textbf{\textit{S}}$, our key point is to transfer the feature from target domain $\textbf{\textit{T}}$ to source domain $\textbf{\textit{S}}$. Thereafter, the depth estimation network should decode the target image's features to depth output. 

Specifically, considering all VIS images are in the same domain, we need only transfer TIR-part features to the VIS domain. For the source VIS image pair $(\boldsymbol{I}^{l}_{S},\boldsymbol{I}^{r}_{S})$ and target TIR-VIS image pair $(\boldsymbol{I}^{l}_{T},\boldsymbol{I}^{r}_{T})$, the VIS encoder $E_{V}$ and the TIR encoder $E_{T}$ encode them to their corresponding features. For adversarial feature transfer, the TIR encoder acts as a generator to generate features $\boldsymbol{f}^{l}_{T}$ from the input TIR image $\boldsymbol{I}^{l}_{T}$, and the VIS encoder encodes $\boldsymbol{f}^{l}_{S}$ from $\boldsymbol{I}^{l}_{S}$ which acts as a ``real feature". The discriminator $D$ try to distinguish whether $\boldsymbol{f}^{l}_{T}$ is a ``real" or ``fake" feature that encoded from $\textbf{\textit{T}}$. Thus, the TIR encoder is trained to reduce the distance between the distributions of TIR and VIS features and encode VIS-like features from a TIR image. The loss for adversarial feature transfer is 
\begin{align}
    \begin{split}
        L_{GAN}({E}_{T},D) & = \min _{E_T}\max _{D}V({E}_{T},D) \\
        & = E_{\boldsymbol{f}_{S}\sim p(\boldsymbol{f}_{S})}[log(D(\boldsymbol{f}_{S}))] \\
        & + E_{\boldsymbol{f}_{T}\sim p(\boldsymbol{f}_{T})}[log(1-D({E}_{T}(I_T)))].
    \end{split}
\end{align}

Since features from the encoder network have different scales due to the varying convolutional layers, we train multiple discriminators instead of a single discriminator network. For the feature $f_{i}, i=1,2,...,L$ of each scale, we use the corresponding discriminator $D_{i}, i=1,2,...,L$ to process domain adaptation step as shown in Figure \ref{fig:fullnet}. 

Because our cross-spectrum feature transfer task belongs to the unpaired domain transfer problem, the generator ($E_{T}$ in our network) only translates data of the domain $\textbf{\textit{T}}$ to match the data distribution of the domain $\textbf{\textit{S}}$, and cannot translate the data to a specific paired one in the domain $\textbf{\textit{S}}$. 
In practice, it often leads the network to transfer all input data to the same output and lose its information of the input data. Therefore, adversarial learning only transfers the data distribution to the same domain and could not get accurate depth output. Inspired by \cite{zhu2017unpaired}, we add a feature reconstruction network to solve this problem. 

Based on the output feature $\boldsymbol{f}^{l}_{T}$ and $\boldsymbol{f}^{r}_{T}$, we train two reconstruction networks, a VIS reconstruction $R_{V}$ and a TIR reconstruction $R_{T}$, to reconstruct features to their corresponding input TIR or VIS image, respectively. The reconstruction process ensures that the encoder will keep the input image's information. We minimize it with the L1 loss as follows,
\begin{align}
    \begin{split}
        L_{rec} &= ||R_{T}(\boldsymbol{f}^{l}_{T})-\boldsymbol{I}^{l}_{T} ||_{1} \\
         & + ||R_{V}(\boldsymbol{f}^{r}_{T})-\boldsymbol{I}^{r}_{T} ||_{1}.
    \end{split}
\end{align}

Meanwhile, as the features are transferred to the same domain, we can input TIR features $\boldsymbol{f}^{l}_{T}$ to $R_{V}$, and it should output a VIS image of a right view. This ``right VIS image" should be the same as the original left VIS image warped to the right view with the corresponding depth $\boldsymbol{d}^{r}$. Also, applying the same reconstruction operation to the VIS features, we can get the new warp reconstruction loss as 
\begin{align}
    \begin{split}
        L_{wrec} &=  ||R_{T}(\boldsymbol{f}^{r}_{T})-\omega(\boldsymbol{I}^{r}_{T},\boldsymbol{d}^{l}) ||_{1} \\
         & + ||R_{V}(\boldsymbol{f}^{l}_{T})-\omega(\boldsymbol{I}^{l}_{T},\boldsymbol{d}^{r}) ||_{1},
    \end{split}
\end{align}
where $\omega()$ is a warping operation given depth. 
With these reconstruction steps, the network can be optimized to output the correct depth and reconstructed image together. Combining two reconstruction loss and adversarial transfer loss, the full loss for feature domain transfer is 
\begin{align}
    L_{tr} = L_{GAN}({E}_{T},D) + L_{rec}+L_{wrec}.
\end{align}

In each training step, we first train the unsupervised (domain $\textbf{\textit{S}}$) depth estimation part based on VIS image pairs. Then we transfer the TIR-VIS image pair (domain $\textbf{\textit{T}}$) by domain feature transfer. Feeding transferred features to depth decoder, we get the dual-spectrum depth output.

Since the TIR-VIS image pair is also a stereo, similar to the VIS unsupervised depth estimation, we also use left-right consistency loss $L_{lr}$ and depth smoothness loss $L_{s}$ to optimize stereo depth estimation. The final loss of cross-spectrum depth is 
\begin{align}
    \begin{split}
        L_{ms} &= L_{tr} + \lambda _{s}(L_{s}(\boldsymbol{I}^{l}_{T})+L_{s}(\boldsymbol{I}^{r}_{T}))\\
                &+\lambda _{lr}(L_{lr}(\boldsymbol{I}^{l}_{T})+L_{lr}(\boldsymbol{I}^{r}_{T})).
    \end{split}
\end{align}

\begin{table*}[ht]
    \centering
    \begin{threeparttable}
    \begin{tabular}{c c c c c c}
        \toprule
         \textbf{dataset} & \textbf{camera} & \textbf{spectral type} & \textbf{data number} & \textbf{label} \\
        \midrule
         CMU\cite{zhi2018deep} & stereo & NIR-VIS & 11k & manual labeled\\
         KAIST\cite{kim2018multispectral} & stereo(one optical aligned camera) & TIR-VIS & 7383 & lidar point\\
         NUST-SR\cite{li2020ivfusenet} & mono aligned & TIR-VIS & 11k & lidar point \\
         VTD(ours) & stereo(two optical aligned camera) & TIR-VIS & 10.5k & lidar point\\
        \bottomrule
    \end{tabular}
    \end{threeparttable}
    \caption{Comparison between the existing datasets and our VTD dataset. }
    \label{tab:tabel1}
\end{table*}

\begin{figure*}[ht]
    \centering
    \includegraphics[width=0.9\textwidth]{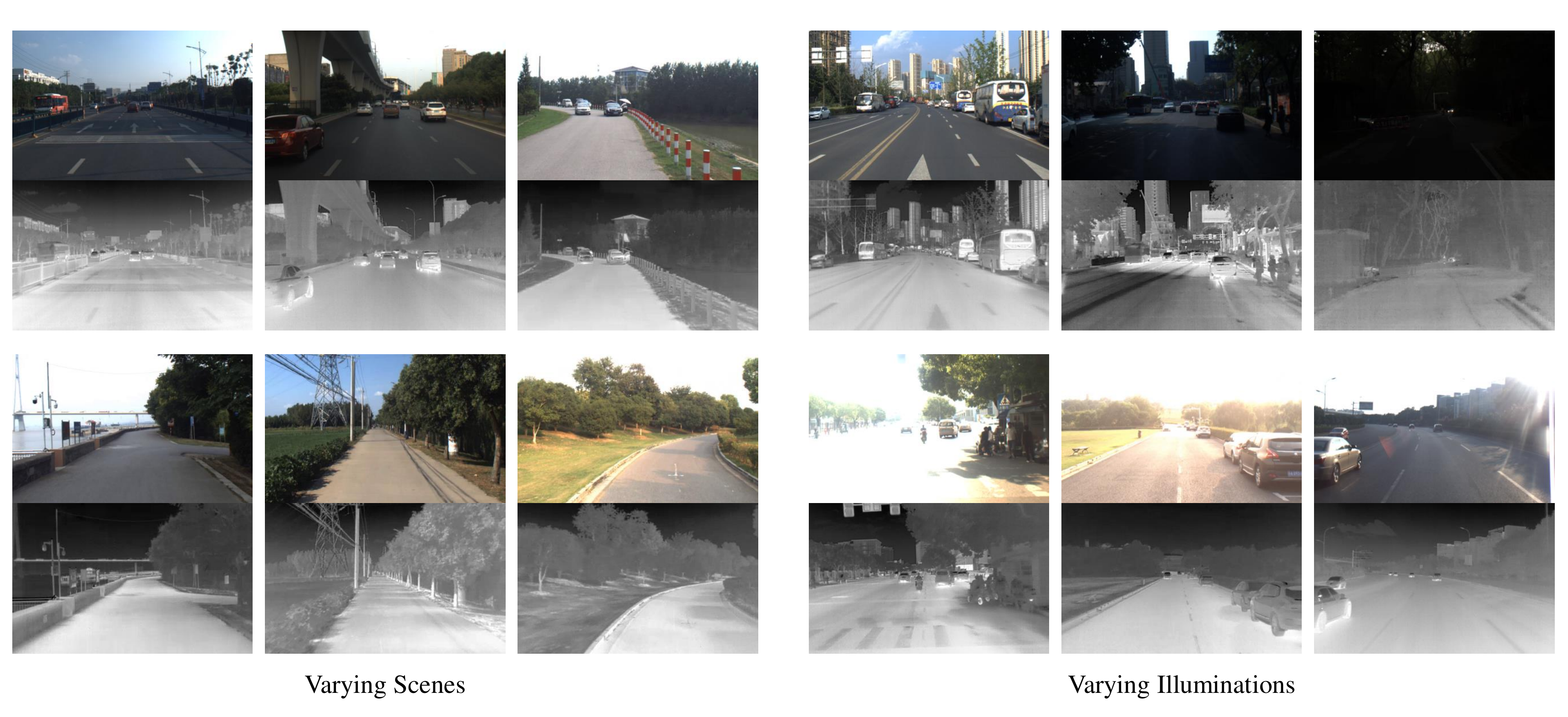}
    \caption{Image examples show the TIR-VIS stereo image pair. The left three columns show images captured in different scenes. The right three columns show images captured under different illumination conditions.}
    \label{fig:sence}
\end{figure*}

\subsection{Cross-spectrum Depth Cycle Consistency}
Since VIS-TIR stereo images also follow geometry correspondence relation between the output depth and the input image, we can find some geometry transformation consistency between them with a cycle loop for a further depth estimation, which is called the cross-spectrum depth cycle consistency.


As shown in Figure \ref{fig:fullnet}, given the TIR-VIS stereo image pair $(\boldsymbol{T}_{L},\boldsymbol{V}_{R})$ and their depth output $\boldsymbol{d}^{l},\boldsymbol{d}^{r}$, we can warp $\boldsymbol{T}_{L}$ with $\boldsymbol{d}^{r}$ to get $\tilde{\boldsymbol{T}_{R}}$, where $\tilde{\boldsymbol{T}_{R}}$ should be a right-view image with original spectral of $\boldsymbol{T}_{L}$. Getting $\tilde{\boldsymbol{V}_{L}}$ by the same way, we get a new VIS-TIR stereo image pair $(\tilde{\boldsymbol{V}_{L}},\tilde{\boldsymbol{T}_{R}})$ that exchange spectral of left and right view image. Sending this new image pair to depth estimation network to get new output depth $\hat{\boldsymbol{d}^{l}},\hat{\boldsymbol{d}^{r}}$, and new output depth should same to original depth $\boldsymbol{d}^{l},\boldsymbol{d}^{r}$. With cross-spectral depth cycle consistency, the image warped by output depth will keep its semantic information consist and satisfy the transformation of stereo geometric reprojection. The loss of cross-spectral depth cycle consistency is
\begin{align}
    L_{cyc} = ||\boldsymbol{d}^{l}-\hat{\boldsymbol{d}^{l}}||_{1} + ||\boldsymbol{d}^{r}-\hat{\boldsymbol{d}^{r}}||_{1}.
\end{align}

Combining all above parts, the final total loss for the full framework is
\begin{align}
    \begin{split}
        L &= \lambda _{v}L_{v} + \lambda _{ms} L_{ms} + \lambda _{cyc} L_{cyc}.
    \end{split}
\end{align}

\section{Experiments}

\begin{table*}[ht]
    \centering
    \begin{threeparttable}
    \begin{tabular}{c c c c c c c c}
        \toprule
         \textbf{Method} & \textbf{abs rel} $\downarrow$ & \textbf{sq rel} $\downarrow$ & \textbf{rmse} $\downarrow$ & \textbf{rmse log}  $\downarrow$ & \textbf{$\Delta<1.25$} $\uparrow$ & \textbf{$\Delta<1.25^2$} $\uparrow$ & \textbf{$\Delta<1.25^3$} $\uparrow$ \\
        \midrule
          monodepth2(RGB) & 0.3022 & 11.3284 & 14.5683 & 0.3415 & 0.7237 & 0.8564 & 0.9161 \\
          monodepth2(IR)  & 0.6436 & 32.4048 & 23.7818 & 0.5712 & 0.5816 & 0.7376 & 0.8196 \\
          DASC            & 2.3313 & 162.5851  & 43.0294 & 1.1249 & 0.2690 & 0.4278 & 0.5261\\
          CSS            & 1.0215 & 44.0977 & 26.004 & 1.0655 & 0.1715 & 0.3325 & 0.4740 \\
          
        \cmidrule(lr){1-8}
          ours(Kitti VIS-guided)  & 0.2889 & 8.0762 & 11.3671 & 0.3617 & 0.7202 & 0.8633 & 0.9280 \\
          ours(VTD VIS-guided)        & \textbf{0.1993} & \textbf{5.4964}  & \textbf{10.2276} & \textbf{0.2747} & \textbf{0.7879} & \textbf{0.9134} & \textbf{0.9593} \\
        
        \bottomrule
        
    \end{tabular}
    \end{threeparttable}
    \caption{Comparison result with other methods. Metric evaluation of our and other methods for  cross-spectrum depth estimation on our dataset. $\downarrow$ means the lower the better and $\uparrow$ means the higher the better.} 

    \label{tab:table2}
\end{table*}

\begin{figure*}
    \centering
    \begin{subfigure}[b]{0.95in}
        \includegraphics[width=0.95in]{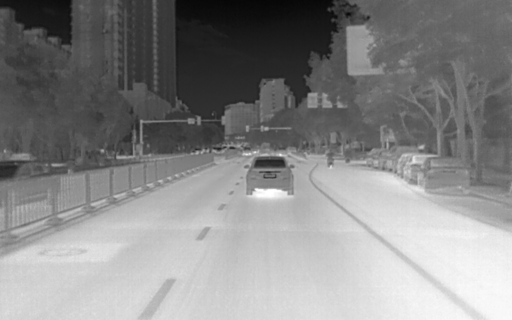}
    \end{subfigure}
    \begin{subfigure}[b]{0.95in}
        \includegraphics[width=0.95in]{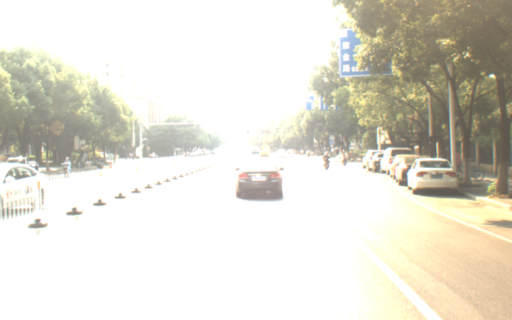}
    \end{subfigure}
    \begin{subfigure}[b]{0.95in}
        \includegraphics[width=0.95in]{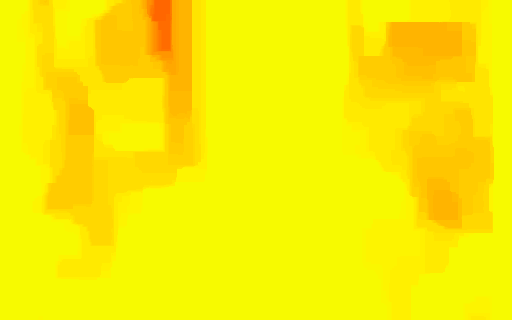}
    \end{subfigure}
    \begin{subfigure}[b]{0.95in}
        \includegraphics[width=0.95in]{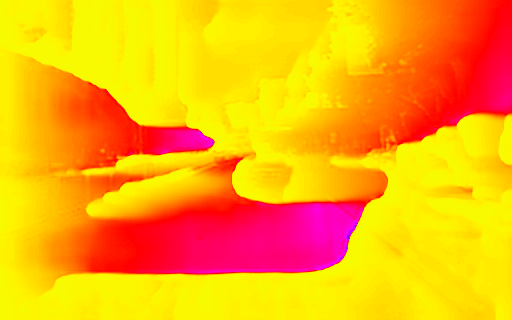}
    \end{subfigure}
    \begin{subfigure}[b]{0.95in}
        \includegraphics[width=0.95in]{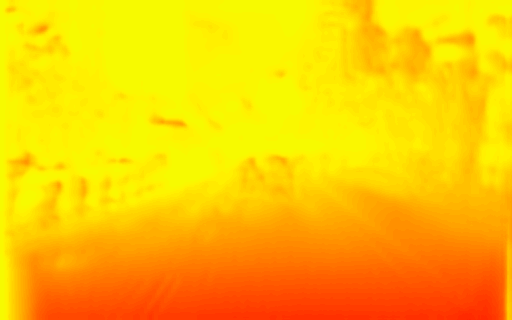}
    \end{subfigure}
    \begin{subfigure}[b]{0.95in}
        \includegraphics[width=0.95in]{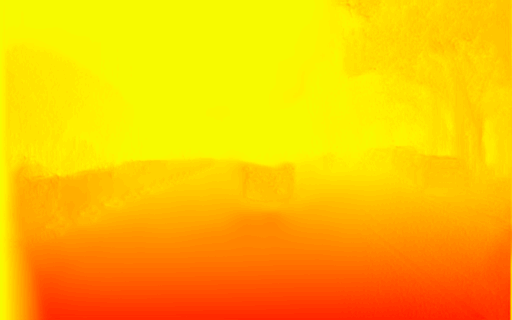}
    \end{subfigure}
    \begin{subfigure}[b]{0.95in}
        \includegraphics[width=0.95in]{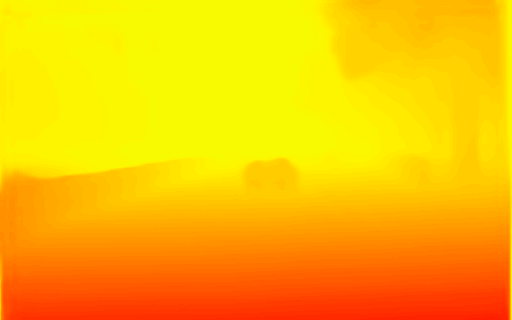}
    \end{subfigure}
    \vspace{5pt}
    
    \begin{subfigure}[b]{0.95in}
        \includegraphics[width=0.95in]{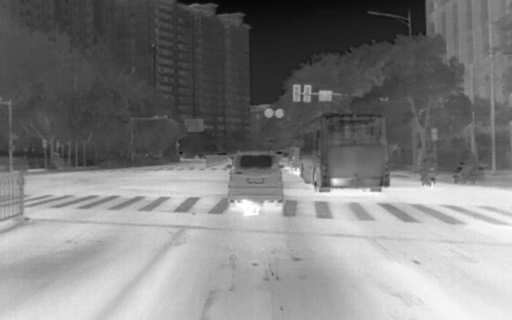}
    \end{subfigure}
    \begin{subfigure}[b]{0.95in}
        \includegraphics[width=0.95in]{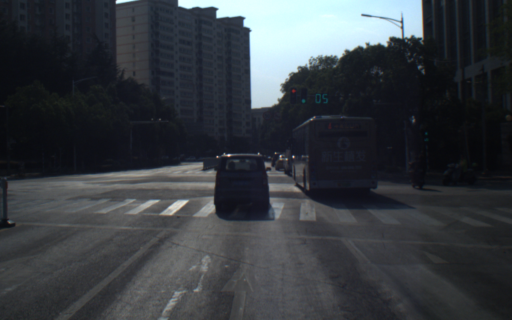}
    \end{subfigure}
    \begin{subfigure}[b]{0.95in}
        \includegraphics[width=0.95in]{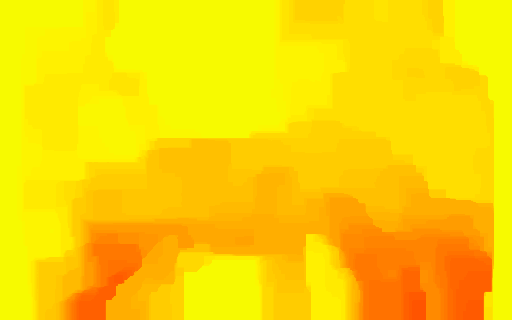}
    \end{subfigure}
    \begin{subfigure}[b]{0.95in}
        \includegraphics[width=0.95in]{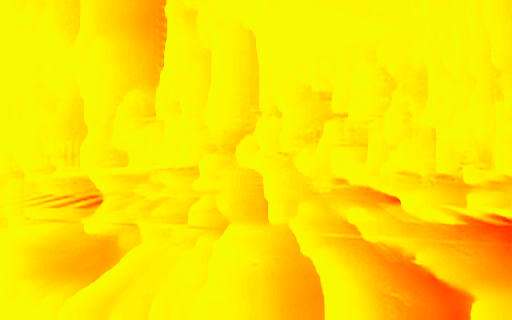}
    \end{subfigure}
    \begin{subfigure}[b]{0.95in}
        \includegraphics[width=0.95in]{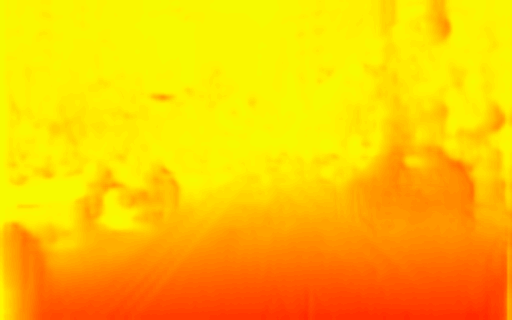}
    \end{subfigure}
    \begin{subfigure}[b]{0.95in}
        \includegraphics[width=0.95in]{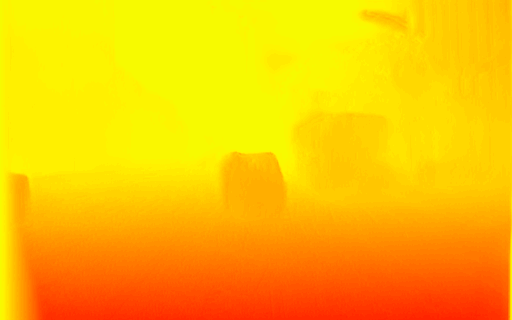}
    \end{subfigure}
    \begin{subfigure}[b]{0.95in}
        \includegraphics[width=0.95in]{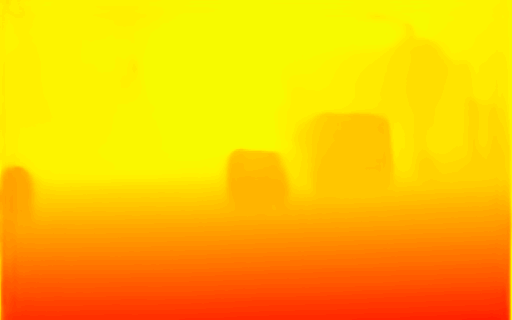}
    \end{subfigure}
    \vspace{5pt}
    
    \begin{subfigure}[b]{0.95in}
        \includegraphics[width=0.95in]{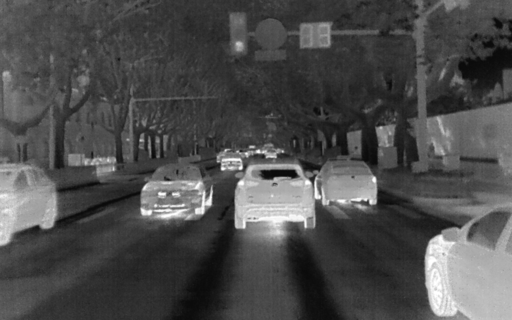}
    \end{subfigure}
    \begin{subfigure}[b]{0.95in}
        \includegraphics[width=0.95in]{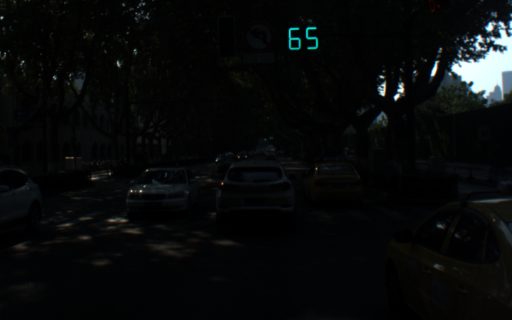}
    \end{subfigure}
    \begin{subfigure}[b]{0.95in}
        \includegraphics[width=0.95in]{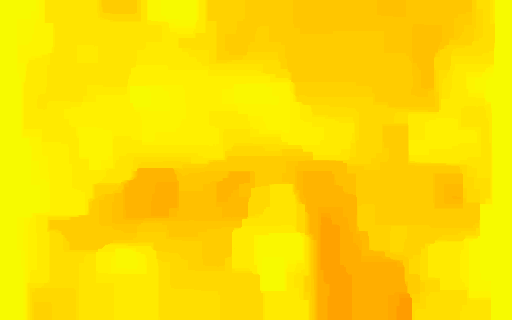}
    \end{subfigure}
    \begin{subfigure}[b]{0.95in}
        \includegraphics[width=0.95in]{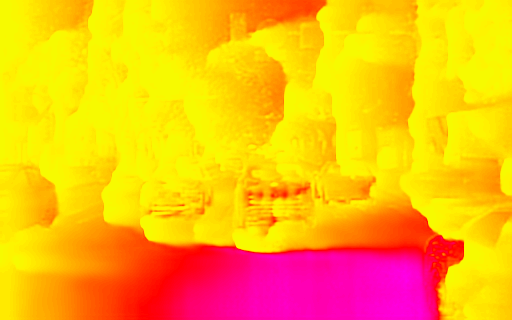}
    \end{subfigure}
    \begin{subfigure}[b]{0.95in}
        \includegraphics[width=0.95in]{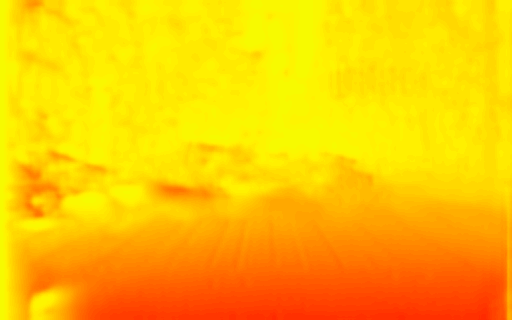}
    \end{subfigure}
    \begin{subfigure}[b]{0.95in}
        \includegraphics[width=0.95in]{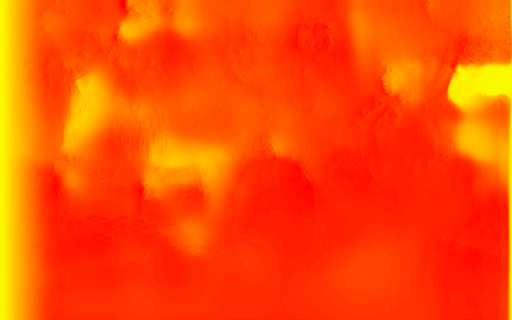}
    \end{subfigure}
    \begin{subfigure}[b]{0.95in}
        \includegraphics[width=0.95in]{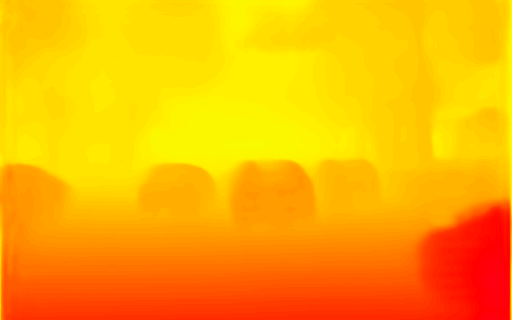}
    \end{subfigure}
    \vspace{5pt}
    
    \begin{subfigure}[b]{0.95in}
        \includegraphics[width=0.95in]{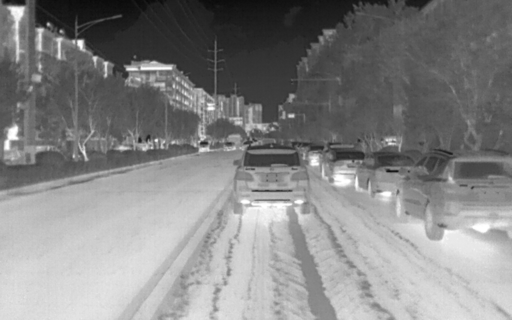}
        \caption{left TIR}
    \end{subfigure}
    \begin{subfigure}[b]{0.95in}
        \includegraphics[width=0.95in]{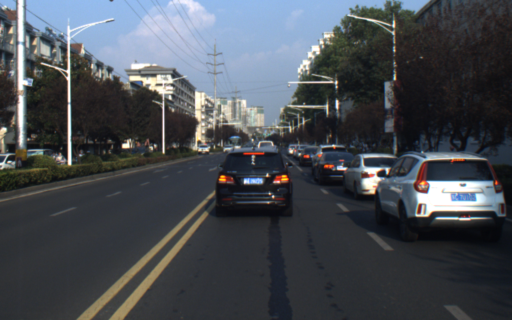}
        \caption{right VIS}
    \end{subfigure}
    \begin{subfigure}[b]{0.95in}
        \includegraphics[width=0.95in]{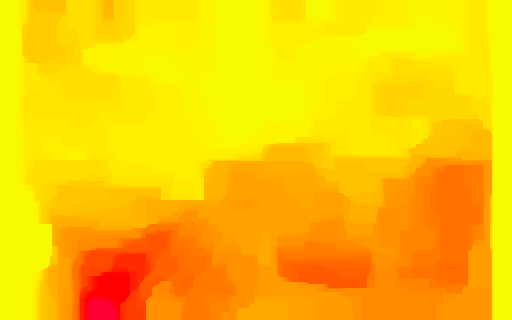}
        \caption{DASC}
    \end{subfigure}
    \begin{subfigure}[b]{0.95in}
        \includegraphics[width=0.95in]{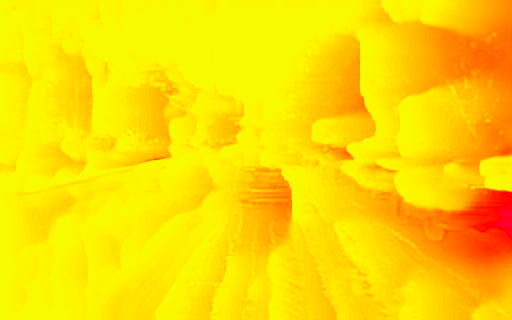}
        \caption{CSS}
    \end{subfigure}
    \begin{subfigure}[b]{0.95in}
        \includegraphics[width=0.95in]{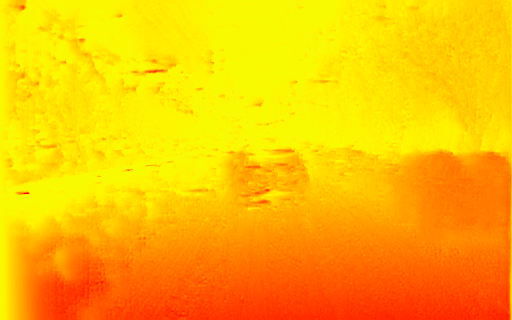}
        \caption{mono2 TIR}
    \end{subfigure}
    \begin{subfigure}[b]{0.95in}
        \includegraphics[width=0.95in]{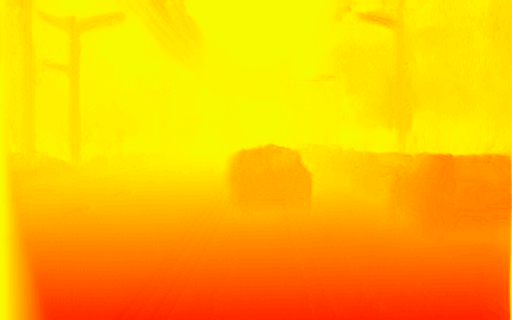}
        \caption{mono2 VIS}
    \end{subfigure}
    \begin{subfigure}[b]{0.95in}
        \includegraphics[width=0.95in]{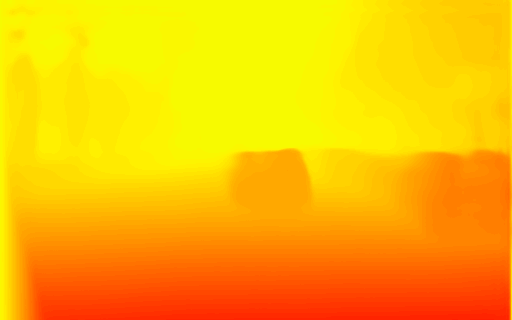}
        \caption{ours}
    \end{subfigure}
    \vspace{5pt}
    
    \caption{Comparisons experiments result: (a) the TIR image, (b) the VIS image, (c) DASC \cite{kim2015dasc},(d) CSS \cite{liang2019unsupervised}, (e) monodepth2(stereo) \cite{godard2019digging} with TIR input, (f) monodepth2(stereo) with VIS input, (g) our method(VTD VIS guided).}
    \label{fig:comparedresult}
\end{figure*}

\subsection{Dataset and Evaluation Baseline}
Due to the lack of open-source large datasets for cross-spectrum depth estimation, we release a large VIS-TIR dataset (VTD) for outdoor road scenes. This dataset is created based on our previously developed hybrid RGB-Thermal cameras \cite{zhang2019build,zhang2022build}, which can output pixel-wisely aligned RGB and thermal images based on a beam-splitter. To create our VTD dataset, we use two such hybrid cameras (C$_h^{L}$ and C$_h^{R}$). The hybrid camera C$_h^{L}$ can obtain a pixel-wisely aligned thermal image $T_{L}$ and RGB image $V_{L}$. Likewise, the hybrid camera C$_h^{R}$ can obtain a pixel-wisely aligned thermal image $T_{R}$ and RGB image $V_{R}$. Our dataset is created based on the thermal images $T_{L}s$ and RGB images $V_{R}s$. Of course, we can also use thermal images $T_{R}s$ and RGB image $V_{L}s$ to create the VTD dataset. 
The baseline between C$_h^{L}$ and C$_h^{R}$ is 50cm. We also use a Velodyne-64E LiDAR sensor to obtain accurate 3D depth measurement as a ground-truth for the TIR-VIS stereo depth estimation, which can be achieved based on a calibration of relative position between the LiDAR and hybrid cameras. 


The image resolution is $1280 \times 800$ and the frame rate of both VIS and TIR cameras is 30Hz. We collected our dual-spectrum image data within a variety of urban road scenes and under illumination conditions. To avoid excessive repetition between image data, we extract images from the original data frame at approximately one-second intervals to compose the final dataset. Totally, we get about 10.5k images in the daytime on different roads. 
Data collection locations include the city center, suburbs, parks and hill trails. We collect data all day from 9am to 6pm to guarantee data collection under different illumination conditions. Figure \ref{fig:sence} shows the example of our data. Compared to the other existing datasets (Table \ref{tab:tabel1}), our dataset has more diverse data and reliable ground truth for training and evaluation. More details about our dataset can be found in the supplementary material.

For evaluation experiments, about 80 percent of the data are used as training sets randomly and the others for testing sets. To evaluate the depth estimation result, the ground truth depth measured by lidar is used.  Evaluation metrics are the same as \cite{godard2017unsupervised}. We use four error evaluation metrics: absolute relative error(\textbf{abs rel}), square relative error(\textbf{sq rel}), root mean square error(\textbf{rmse}), root mean square error of log(\textbf{rmse log}), and an accuracy evaluation metrics: depth ratio accuracy(\textbf{$\Delta<1.25$, $\Delta<1.25^2$, $\Delta<1.25^3$}).

\subsection{Network Details}
In our depth estimation net branch, the network structure of the encoder and decoder is the same as the depth net in \cite{godard2019digging}, and we chose resnet18 \cite{he2016deep} for the encoder in our experiments. The discriminator network part contains five discriminator networks following \cite{zhu2017unpaired} with different downsampling numbers for multi-scale features. All encoders are initialized by the weights pre-trained with the ImageNet and the other networks are initialized by a Gaussian distribution with 0 mean and 0.02 standard deviation.
During training step, we set weight parameters of loss as follows: $\lambda _{s}=0.1$, $\lambda _{lr}=1$, $\lambda _{v}=1$, $\lambda _{ms}=1$ and $\lambda _{cyc}=10$. The training data are VIS and TIR images in our released dataset. We use random sampled VIS image pair in our VTD dataset as domain $\textbf{\textit{S}}$ for guided and VIS-TIR image pair in our dataset as domain $\textbf{\textit{T}}$. We update our network by Adam optimizer \cite{kingma2014adam} with 1 batchsize and 0.0002 learning rate. We train the network for 60 epochs on an RTX 2080ti machine. All input image are resized to $512 \times 320$.

\subsection{Comparison Experiments}

We compare our method with a couple of existing methods on our dataset, including monodepth2(stereo) \cite{godard2019digging}, DASC \cite{kim2015dasc} and cross-spectral stereo matching by synthesize (CSS) \cite{liang2019unsupervised}. We report both metric evaluation and depth image visualization results. Table \ref{tab:table2} shows the detailed evaluation result and Figure \ref{fig:comparedresult} shows the image result example for comparison purposes.

As a traditional method, the DASC does not need training data and can be tested directly with VIS-TIR image pair. The CSS transfers two spectral images to each other in image level to train depth estimation network. These two methods mostly fail due to the huge difference between VIS and TIR images.

For monodepth2(stereo), we present the results that are trained with VIS and TIR data image stereo pairs, respectively.  From the experiment result, we can see that the monodepth2(stereo) result with TIR training data provides depth estimation with over smoothness. That is mainly because TIR images have poor texture compared to VIS images and the photometric re-projection loss cannot obtain a stable and accurate result. The Monodepth2(stereo) with VIS images gets high-quality depth results in the scenes with good lighting conditions but fails in some adverse lighting conditions such as the heavy shadow.

Our VIS-guided depth estimation method gets the best evaluation results as shown in Table \ref{tab:table2} because we have taken advantages of both TIR and VIS images in our framework. We can see from figure \ref{fig:comparedresult} that our method can provide a good depth result under varying illumination conditions such as in the dark or overexposed.

Furthermore, we add an experiment that uses VIS image pair data in KITTI \cite{geiger2012we} as images in the domain $\textbf{\textit{S}}$ and our VIS-TIR pair as images in the domain $\textbf{\textit{T}}$ to show the generality of our method in domain feature transfer. Our method also achieves a good result but slightly worse than the original method due to the difference of data distribution between KITTI and our dataset. We denote this experiment as ours(Kitti VIS-guided) and the evaluation result is also shown in Table \ref{tab:table2} bottom.

It deserves to point out that the algorithm in \cite{kim2018multispectral} can achieve better result for cross-spectrum depth estimation than ours. But this method needs a coaxial optical alignment of thermal and visible-light cameras to achieve the pixel-wise alignment of the TIR-VIS image pairs during training. Therefore, this method cannot be generalized to the scenario where the thermal and visible-light cameras are in a general configuration. Thus, we did not perform the comparison experiment.

\subsection{Ablation Study}
We test our method with/without different additional loss configurations to verify their effectiveness based on a basic model. The ablation result is shown in Table \ref{tab:table3}.
\subsubsection{Basic Model}
For our VIS-guided dual-spectrum framework, the basic model only contains the VIS-guided step ($L_{v}$) and feature transfer part ($L_{tr}$). The transfer loss mainly constrains the network encoder to transfer at the feature level to adapt depth output and cannot strictly maintain geometric invariance for the input image. The depth result of the basic model is slightly rough.
\subsubsection{Depth Smoothness and Left-right Consistency}
Similar to the normal depth estimation problem, the depth smoothness loss and depth left-right consistency provide more geometric information for the depth estimation network.  The $L_s$  matches the edge between depth and input image and $L_{lr}$ encourages left/right depth to project to each other. Those two losses both improve the network depth quality.
\subsubsection{Cross-spectrum Depth Cycle Consistency}
Cross-spectrum depth cycle consistency encourages the network to find more geometry transformation constancy between the input images and depth that can keep the image's semantic constant after stereo view warp. From the experiment result, we can find that $L_{cyc}$ significantly improves the accuracy of the depth result.

\begin{table}[ht]
    \centering
    \begin{tabular}{c c c c c}
        \toprule
          & $L_s$ & $L_{lr}$ & $L_{cyc}$ & \textbf{abs rel} $\downarrow$ \\
        \midrule
         Basic Model &  &  &  & 0.5585 \\
         Basic+Depth Smooth & $\surd$ &  &  & 0.4947 \\
         Basic+LR Consistency &  & $\surd$ &  & 0.3755 \\
         Basic+Depth Cycle &  &  & $\surd$ & 0.2340 \\
         Complete Model& $\surd$ & $\surd$ & $\surd$ & 0.1993 \\
        \bottomrule
    \end{tabular}
    \caption{Ablation study with different loss configurations.}

    \label{tab:table3}
\end{table}

\section{Conclusion}
In this paper, we propose a learning-based cross-spectrum depth estimation network without the supervision of ground depth. We introduce a cross-spectrum feature transfer module by a multi-scale feature transfer adversarial network, where the features from TIR are transferred to match the ones from the VIS image and predict correct depth image by the decoder that is trained with VIS image pairs. Furthermore, we combine cross-spectrum depth cycle consistency to improve depth results. Meanwhile, we released a large VIS-TIR depth estimation dataset with lidar measurement as ground truth for evaluation. The experiments on this dataset show that our method has the best performance compared to a few other existing methods. 

\bibliographystyle{IEEEtran}
\bibliography{IEEEabrv,references}

\end{document}